\pdfoutput=1

\documentclass[11pt]{article}

\usepackage[]{ACL2023}

\usepackage{times}
\usepackage{latexsym}

\usepackage[T1]{fontenc}

\usepackage[utf8]{inputenc}

\usepackage{microtype}

\usepackage{inconsolata}

\usepackage{url}
\usepackage{subfigure}
\usepackage{multirow}
\usepackage{enumitem}
\usepackage{stmaryrd}
\usepackage{array}
\usepackage{threeparttable}
\usepackage{blindtext}
\usepackage{amssymb}

\usepackage[ruled,vlined,linesnumbered,longend]{algorithm2e}
\usepackage{svg}
\usepackage{placeins}

\usepackage{soul}
\usepackage[utf8]{inputenc}
\usepackage{graphicx}
\usepackage{amsmath}
\usepackage{booktabs}

\usepackage{lipsum}

\usepackage{listings}
\usepackage{xcolor}
\definecolor{codegreen}{rgb}{0,0.6,0}
\definecolor{codegray}{rgb}{0.5,0.5,0.5}
\definecolor{codepurple}{rgb}{0.58,0,0.82}
\definecolor{backcolour}{rgb}{0.99,0.99,0.98}
\lstdefinestyle{mystyle}{
    backgroundcolor=\color{backcolour},
    commentstyle=\color{codegreen},
    keywordstyle=\color{magenta},
    numberstyle=\tiny\color{codegray},
    stringstyle=\color{codepurple},
    basicstyle=\ttfamily\footnotesize,
    breakatwhitespace=false,
    breaklines=true,
    captionpos=b,
    keepspaces=true,
    numbers=left,
    numbersep=5pt,
    showspaces=false,
    showstringspaces=false,
    showtabs=false,
    tabsize=2,
    frame=shadowbox,
    rulesepcolor=\color{red!20!green!20!blue!20},
    xleftmargin=1em,xrightmargin=0em,aboveskip=1em,
    framexleftmargin=1em,
}

\newcommand{\hide}[1]{}

\newcommand{\eg}{\textit{e.g.,}}

\usepackage{tikz}

\definecolor{parsecolor}{rgb}{0.140625,0.234375,0.5}
\newcommand{\autourParse}[1]{\tikz[baseline=(X.base)]\node [draw=black,fill=parsecolor,semithick,rectangle,inner sep=1.4pt, rounded corners=3pt] (X) {#1};}

\definecolor{runcolor}{rgb}{0.4375,0.203125,0.5078125}
\newcommand{\autourRun}[1]{\tikz[baseline=(X.base)]\node [draw=black,fill=runcolor,semithick,rectangle,inner sep=1.4pt, rounded corners=3pt] (X) {#1};}

\definecolor{addcolor}{rgb}{0.9375,0.9375,0.9375}
\newcommand{\autourAdd}[1]{\tikz[baseline=(X.base)]\node [draw=black,fill=addcolor,semithick,rectangle,inner sep=1.4pt, rounded corners=3pt] (X) {#1};}

\newcommand*{\Parse}{\autourParse{\includegraphics[scale=0.008]{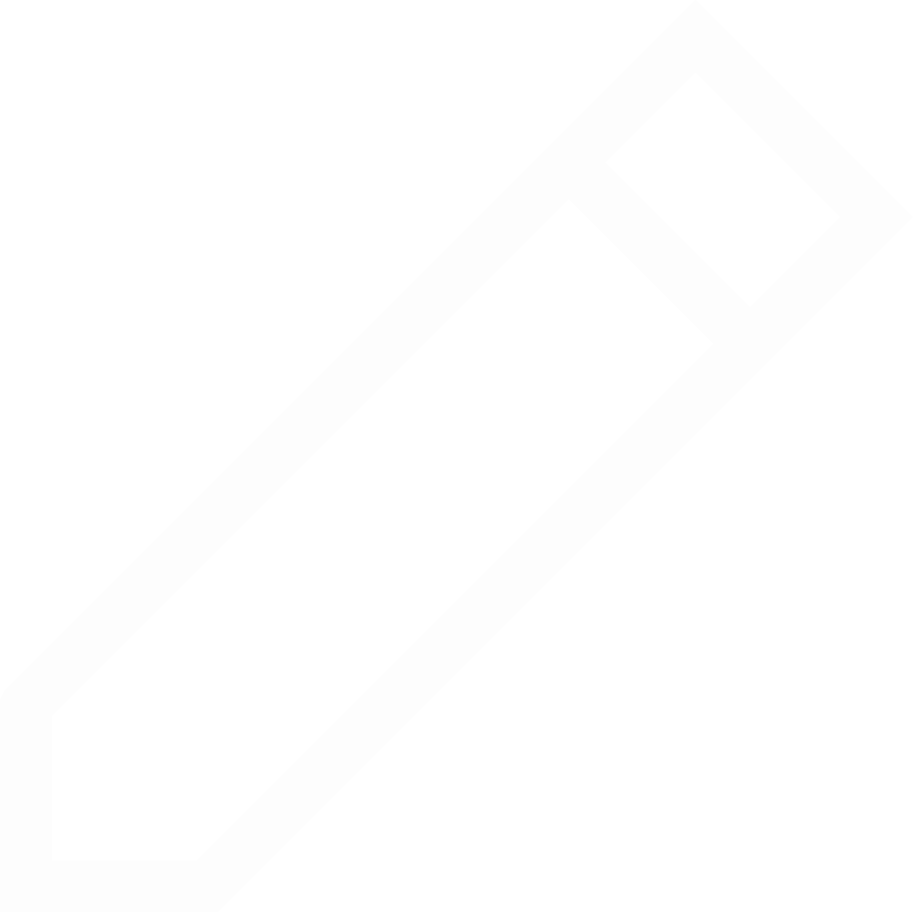}\ \color{white}\texttt{Parse}}\ \ }
\newcommand*{\Run}{\autourRun{\includegraphics[scale=0.007]{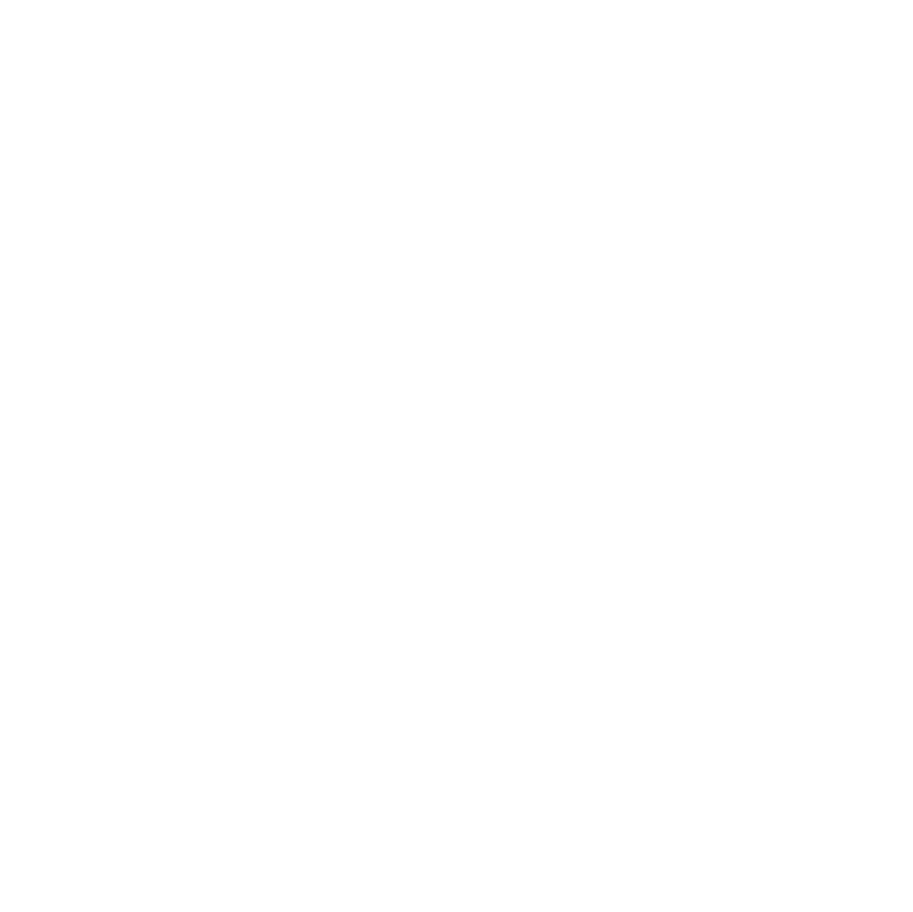}\ \color{white}\texttt{Run}}\ \ }
\newcommand*{\Add}{\autourAdd{\includegraphics[scale=0.45]{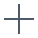}}\ }

\usepackage{amsmath,amsfonts,bm}

\def\eqref#1{equation~\ref{#1}}

\def\1{\bm{1}}

\DeclareMathAlphabet{\mathsfit}{\encodingdefault}{\sfdefault}{m}{sl}
\SetMathAlphabet{\mathsfit}{bold}{\encodingdefault}{\sfdefault}{bx}{n}

\title{
\includegraphics[scale=0.4]{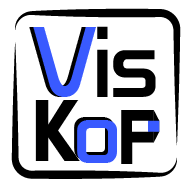} VisKoP: Visual Knowledge oriented Programming \\
for Interactive Knowledge Base Question Answering
}

\newcommand*{\email}[1]{\texttt{#1}}

\author{
Zijun Yao$^{1}$\thanks{\quad Equal contribution.} \quad 
Yuanyong Chen$^{1*}$ \quad
Xin Lv$^{1}$ \quad 
Shulin Cao$^{1}$ \quad 
Amy Xin$^{1}$ \quad
Jifan Yu$^{1}$ \\
{\bf Hailong Jin$^{1}$} \quad 
{\bf Jianjun Xu}$^2$ \quad
{\bf Peng Zhang}$^{1,3}$ \quad
{\bf Lei Hou}$^{1}$\thanks{\quad Corresponding author.} \quad 
{\bf Juanzi Li}$^{1}$ \quad  \\
$^1$Department of Computer Science and Technology, \\
Tsinghua University, Beijing 100084, China \\
$^2$ Beijing Caizhi Technology Co., Ltd. \; $^3$ Zhipu.AI \\
\email{yaozj20@mails.tsinghua.edu.cn, houlei@tsinghua.edu.cn}
}

\begin{document}

\maketitle
\begin{abstract}
We present Visual Knowledge oriented Programming platform (VisKoP), a knowledge base question answering (KBQA) system that integrates human into the loop to edit and debug the knowledge base (KB) queries.
VisKoP not only provides a neural program induction module, which converts natural language questions into knowledge oriented program language (KoPL), but also maps KoPL programs into graphical elements.
KoPL programs can be edited with simple graphical operators, such as \textit{``dragging''} to add knowledge operators and \textit{``slot filling''} to designate operator arguments.
Moreover, VisKoP provides auto-completion for its knowledge base schema and users can easily debug the KoPL program by checking its intermediate results.
To facilitate the practical KBQA on a million-entity-level KB, we design a highly efficient KoPL execution engine for the back-end.
Experiment results show that VisKoP is highly efficient and user interaction can fix a large portion of wrong KoPL programs to acquire the correct answer.
The VisKoP online demo\footnote{\href{https://demoviskop.xlore.cn}{demoviskop.xlore.cn}  (Stable release of this paper) and \href{https://viskop.xlore.cn}{viskop.xlore.cn} (Beta release with new features).\label{foot:viskop}}, highly efficient KoPL engine\footnote{\url{https://pypi.org/project/kopl-engine}\label{foot:engine}}, and screencast video\footnote{\url{https://youtu.be/zAbJtxFPTXo}} are now publicly available.

\end{abstract}

\section{Introduction}


Knowledge Base Question Answering (KBQA) aims to find answers to factoid questions with an external Knowledge Base (KB).
Researchers have fully explored the KBQA~\cite{Lan2021ASO} task and the most common solution is to convert user-posed natural language questions into KB query programs via semantic parsing and then give a final result by executing queries on the KB, such as SPARQL~\cite{mihindukulasooriya2020leveraging,gu2021beyond}, $\lambda$-DCS~\cite{Wang2015BuildingAS,Shin2021ConstrainedLM}, and KoPL~\cite{kqapro2022kopl,cao2022program}.
Recently, many KBQA systems~\citep{hoffner2013tbsl,wanyun2016kbqafreebase,ibrahim2021dtqa,chen2021retrack} that implement those advanced semantic parsing algorithms in an online environment, have been developed.

\begin{figure}[!t]
\centering
    \includegraphics[width=0.98\linewidth]{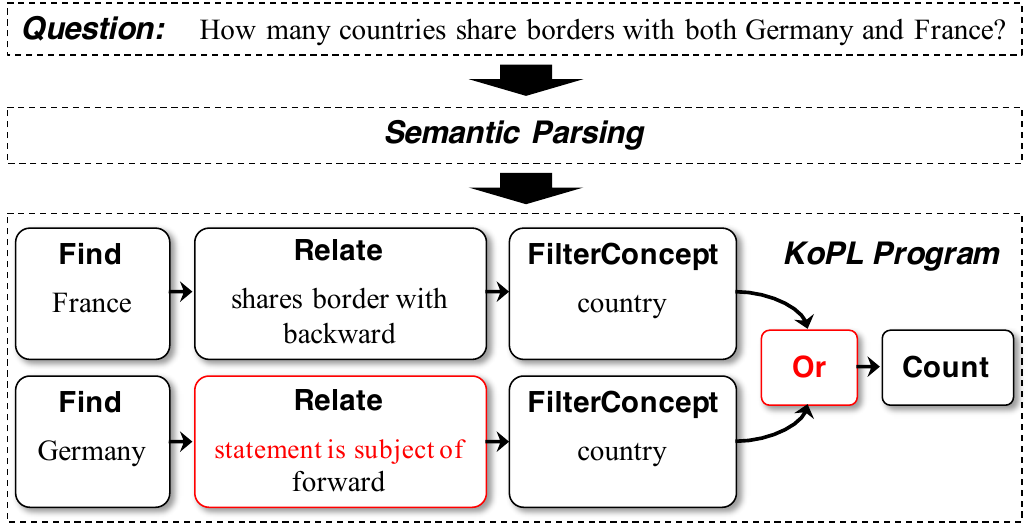}
    \caption{
    Semantic parsing results for natural language question \textit{``How many countries share borders with Germany and France?''} given by state-of-the-art model trained on KQA Pro.
    Errors are marked in {\color{red}red} color.
    }
    \label{fig:example}
\end{figure}

Although semantic parsing methods have gained considerable achievement, there is still no guarantee to precisely parse every user-posed question given the limitations of current machine learning techniques.
Figure~\ref{fig:example} demonstrates an example of semantic parsing results by the state-of-the-art KBQA model~\citep{kqapro2022kopl}.
As the posed question does not follow the identical distribution of the training dataset adopted by the semantic parsing model~\citep{shaw2021compositional,yin2021compositional}, it is falsely parsed with the \textit{Or} operator, which should be an \textit{And} operator, causing the \textbf{structure error} of the KB query.
Meanwhile, it is extremely difficult for the semantic parsing model to correctly predict all the knowledge elements in a question~\citep{cao2022program}.
As shown in the example, the \textit{``shares border with''} relation is falsely predicted as a \textit{``statement is subject of''} relation, causing an \textbf{argument error} in the KB query.
However, existing KBQA systems do not provide easy access to manipulating the KB query programs and thus users cannot intervene in the query execution.



Fortunately, several program-based KB query languages for complex reasoning consisting of modularized operators have come up, making KBQA easy to visualize~\citep{NPI,saha2019complex}.
With applicable visual representation of KB queries, intended users are capable of identifying errors in the programs generated by semantic parsing and correct them.
Based on these observations, we raise a natural question: 
\textit{{
How to design a visualized KBQA system that eases users to inspect and debug those KB query programs?
}}


\textbf{Presented System.}
We demonstrate \textbf{Vis}ual \textbf{K}nowledge \textbf{o}riented \textbf{P}rogramming (VisKoP) platform, an interactive, visualized and program-based KBQA system.
VisKoP provides an interactive knowledge oriented programming environment, allowing users to monitor and debug the KB queries with graphical operators.
In comparison with existing KBQA systems, VisKoP is easier to use due to its following characteristics:


\begin{itemize}[leftmargin=*]
    \item \textbf{Knowledge oriented programming.} 
    VisKoP is the first KBQA system to support Knowledge oriented Programming Language (KoPL)~\citep{kqapro2022kopl}. 
    As a program-based KB query language, KoPL provides modularized program style for users to interact with knowledge elements, within its wide range of knowledge operators.
    Besides, KoPL can be converted into various different KB query languages via GraphQ IR~\cite{graphqir}.
    \item \textbf{Visualized interface.}
    VisKoP maps programming with KoPL into a series of graphical operations---\textit{``dragging''} to add new knowledge operators, \textit{``connecting''} the knowledge operators to add dependencies, and \textit{``slot-filling''} to specify knowledge arguments.
    \item \textbf{Interactive programming and debugging.}
    We use semantic parsing algorithms to convert natural language questions into KB queries, whose execution gives not only the final answers, but also intermediate results of each knowledge operator, which facilitates debugging.
    Meanwhile, auto-completion for KB schema (\eg relation, concept, and attribute) provided by VisKoP assists users that are unfamiliar with the KB schema.
    \item \textbf{High efficiency.}
    We develop a high performing KoPL engine for VisKoP's back-end.
    It executes KoPL on a million-entity level KB in less than $200$ milliseconds, which can hardly be sensed next to the network latency.
\end{itemize}

We conduct user study and find that with the help of the visualized programming interface, users can find the correct answer in an average $110.6$ seconds, which alleviates the problem caused by error-prone semantic parsing algorithms.
Meanwhile, our efficiency study shows that the execution engine is significantly faster than the original KoPL engine and Virtuoso by $16\times$ and $5\times$, respectively. 

\textbf{Contributions.} 
(1) We design a visualized knowledge oriented programming platform for KBQA, which integrates human into the loop to write and debug KB queries.
(2) We implement a high performing KoPL execution engine that scales KoPL to an up-to-date million-entity-level KB.

The development and deployment of VisKoP validates the effectiveness of allowing questioners to monitor the error in the KB queries.
The visual programming platform provides external human guidance on the neural program induction model, and potentially improves the robustness the system.

\section{Preliminaries}


\subsection{Knowledge Base}

As defined by KoPL~\citep{kqapro2022kopl}, KB consists of $4$ kinds of basic knowledge elements:


\noindent\textbf{Entities} are unique objects that are identifiable in the real world, \eg~\textit{Germany}.

\noindent\textbf{Concepts} are sets of entities that have some characteristics in common, \eg~\textit{Country}.

\noindent\textbf{Relations} depict the relationship between entities or concepts. Entities are linked to their concepts via relation \textit{instance of}, while concept  hierarchy is organized via relation \textit{subclass of}.

\noindent\textbf{Attributes} link entities to data value descriptions, \eg~\textit{day of birth}. Attributes can be further classfied into 4 types: date, year, string, and numbers.

These knowledge elements are further organized into $3$ kinds of structured representation in triplets:


\noindent \textbf{Relational knowledge} are triplets organized as $(${head entity, relation, tail entity}$)$.

\noindent \textbf{Literal knowledge} are triplets organized as $(${entity, attribute, value}$)$.
    
\noindent \textbf{Qualifier knowledge} are bound with relational or literal knowledge to specify under which condition they are true. The qualifiers are organized as $(${relational/attribute knowledge, attribute, value}$)$.

\begin{figure*}[!t]
\centering
    \includegraphics[width=0.96\linewidth]{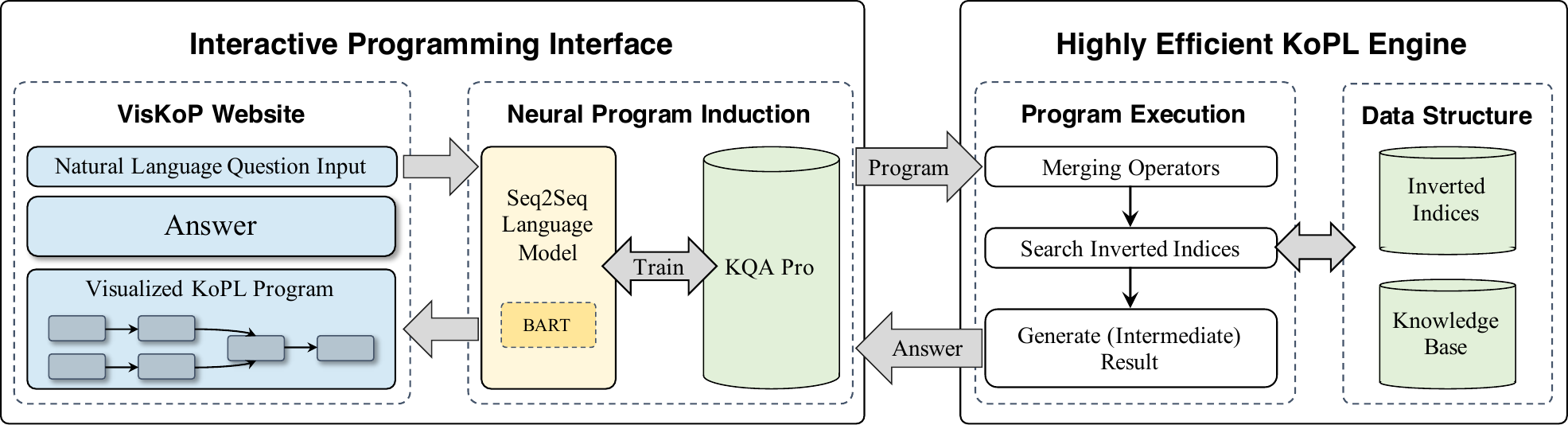}
    \vspace{-0.02in}
    \caption{
    The overall system architecture of VisKoP.
    }
    \label{fig:arch}
\end{figure*}

\subsection{Knowledge Base Question Answering}
KBQA provides a natural-language-based interface for users to access knowledge in the KB. 
It inputs a natural language question $\boldsymbol{q}=\{q_1, \ldots, q_n\}$, where $q_i$ is the $i^{\text{th}}$ word, and outputs the answer utterance $\boldsymbol{a}$.
The answer is either the knowledge elements (\eg~entity name) in the KB, or the result of a combination of logical or algebraic operations performed on the knowledge elements.



\subsection{KoPL}


KoPL stands for knowledge oriented programming language consisting of $26$ different knowledge operators.
Natural language questions can be presented as KoPL programs, which are constructed as connected knowledge operators.
Each operator has two categories of input: operator argument(s), and dependency input(s).
Operator arguments are instructions on how to perform the operator, which are usually determined by the semantics of the question;
Dependency inputs are outputs of previous knowledge operators that are linked to the current operator.
For example, in Figure~\ref{fig:example}, operator \textit{Relate(shares border with, forward)} has two arguments---\textit{shares border with} and \textit{forward}, while the dependency input comes from the \textit{Find} operator.
KoPL programs can be executed on the background KB to obtain the answer. More details are included in Appendix~\ref{sec:app:kopl}.

One essential characteristic of KoPL is that, as modularized knowledge operators, the intermediate result of each operator is preserved and can thus be inspected and debugged.
Given the modularity and inspectability of KoPL, we design the VisKoP platform, as described below.




\section{The VisKoP Platform}

The implementation of our VisKoP platform focuses on $4$ designing principles:

\textbf{I. {Graphical Element Visualization:}} User-posed questions should be parsed into the KoPL program, and shown as graphical elements.

\textbf{II. {Interactive Programming:}} The system needs to enable users to edit and correct the KoPL program with knowledge schema auto-completion and intermediate results demonstration.

\textbf{III. {Highly Efficient Execution:}}  The system should support large scale KBs for practical usage with low execution latency.

\textbf{IV. {Transparent Execution:}} The execution footprint of each operator should be preserved for inspection within interactive programming.

In particular, the first two principles are undertaken by the interactive programming interface in the front-end and the last two principles are undertaken by the highly efficient KoPL program execution engine in the back-end.
The overall architecture of VisKoP is shown in Figure~\ref{fig:arch}.

The implemented VisKoP is deployed as an openly available website\textsuperscript{\ref{foot:viskop}}.
The highly efficient KoPL execution engine is also provided as an open-source Python extension toolkit\textsuperscript{\ref{foot:engine}}.

\subsection{Interactive Programming Interface}



\textbf{Graphical Element Visualization.}
VisKoP allows users to ask natural language questions and parse them into KoPL programs instead of writing KoPL programs from scratch.
The process is carried out by a neural program induction module, as shown in Figure~\ref{fig:arch}, whose backbone is a sequence-to-sequence pre-trained language model.
Here we choose BART~\citep{lewis2020bart} as the backbone and fine-tune it on the KQA Pro dataset~\citep{kqapro2022kopl}. 
It accepts natural language questions as input, and output the KoPL program in the depth first search order. 
The KoPL programs are converted to meet the format of sequence generation.



VisKoP visualizes KoPL program as a tree structure in the editing panel, where the nodes in the tree are knowledge operators with arguments.
Argument inputs are modeled as filling slots in the knowledge operators and dependency inputs are modeled as directed edges between different knowledge operators.
We define 3 kinds of graphical actions that users may take within the KoPL program:
\textit{dragging} to add new operators, \textit{linking} to indicate knowledge elements flow, and \textit{slot-filling} to designate arguments of the knowledge operators.

\begin{figure*}[!t]
\centering
\includegraphics[width=0.97\linewidth]{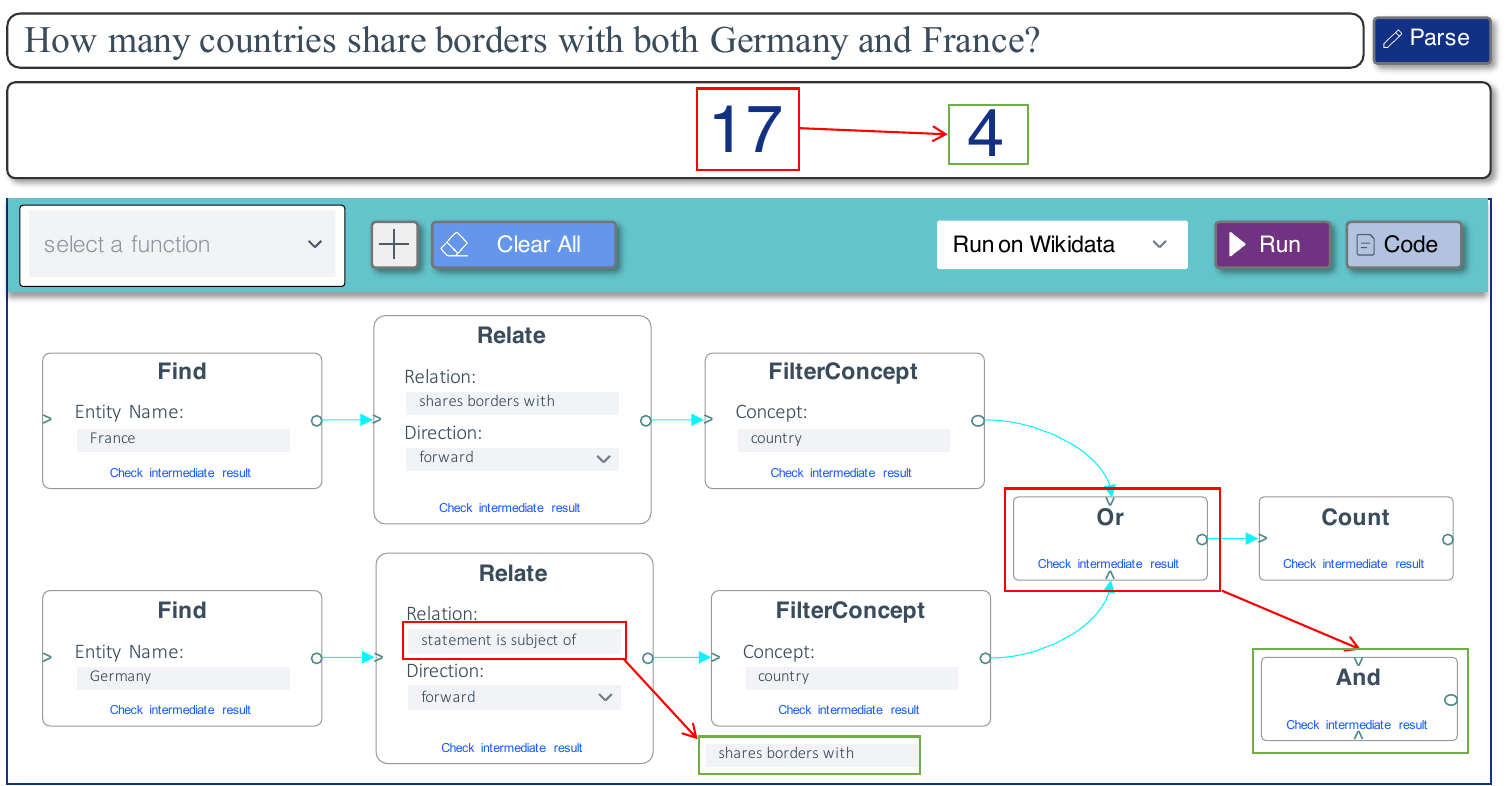}
\vspace{-0.05in}
    \caption{
    Screenshot of the interactive programming interface of VisKoP.
    When user tries to parse ``How many countries share borders with both Germany and France?'', the semantic parsing algorithm falsely predict the \textit{Or} operator, and one of the argument inputs of the \textit{Relate} operator.
    This further results in the wrong answer \textit{``17''}.
    We marked this errors in the red box, and put the correct graphical elements in the nearby green box.
    }
    \label{fig:screen}
\end{figure*}

\textbf{Interactive Programming.}
For users that are less skilled at KoPL programming or less familiar with the schema of the underlying KB, VisKoP implements a series of auxiliary functions.
Firstly, the KB schema is mainly associated with arguments of the knowledge operators.
VisKoP helps to auto-complete knowledge elements via string matching when users try to fill in the argument slots.
Next, to ensure the grammatical correctness of the KoPL program whose users submit to run, we implement linking legitimacy checking.
VisKoP warns users when the the submitted program is not a tree or the dependency is illegal (\eg~The output of the \textit{Count} operator cannot be input to the \textit{QFilterStr} operator).
Finally, intermediate execution results of each knowledge operator are returned from the back-end and presented on the visualized interface where users may debug their KoPL program.

\subsection{Highly Efficient KoPL Engine}

The highly efficient KoPL engine is responsible for most parts of the back-end by reading the KoPL program as input and outputing the answer.


\textbf{Highly Efficient Execution.}
KoPL program execution should be highly efficient for supporting large-scale KBs.
Towards this goal, we adopt three implementation strategies: inverted indices construction, knowledge operators merging, and data structure optimization.

The first step is to construct inverted indices, which maps different types of attribute values and relations to their involved entities.
These inverted indices prevent knowledge operators from enumerating over all the entities in the KB to recall corresponding knowledge elements.
Subsequently, the great deal of time consumed by the engine to filter out entities satisfying certain constraint from the overall KB comes to our attention.
This is represented by consecutive \textit{FindAll} operator and filtering operators (\eg\ \textit{FilterStr}).
We propose to merge the two consecutive operators and construct corresponding inverted indices.
Finally, for all key-value pair data structures, we use the running time of the questions in the KQA Pro dataset on the million-entity level KB as the metric, to greedily search out the optimal storage structure.
The searching space contains hash map, red-black tree, trie tree, and ternary search tree.

\textbf{Transparent Execution.}
Showing the intermediate results in the front-end requires the execution engine to preserve the outputs of each operator in the KoPL program and use them to monitor the behavior of the knowledge query.
Meanwhile, users can debug the input KoPL program by inspecting the intermediate results to locate the bug.

\section{Usage Example}

\subsection{Interactive Programming Interface}

The online website of VisKoP is illustrated in Figure~\ref{fig:screen}.
We give an example of how to interact with the system to obtain the correct answer by questioning \textit{``How many countries share borders with both Germany and France?''}, which cannot be correctly parsed by the semantic parsing algorithm.

\textbf{Neural program induction.}
VisKoP accepts KB queries in natural language. 
The users input the question in the input box on the top of the website.
Clicking on the \Parse button parses the natural language question into its corresponding KoPL program, to be displayed on the editing panel at the bottom of the website. 
The predicted answer is shown by clicking the \Run button in the top of the editing panel.
Here, VisKoP provides the common functionality as a KBQA system.

\textbf{KoPL program debugging.}
As shown by Figure~\ref{fig:screen}, users can easily identify two errors.
One issue comes from the structural aspect.
The answer should be counted on the intersection of two sets of country, each sharing border with Germany and France, respectively.
To replace the operator \textit{Or} with the operator \textit{And}, users may first click on the \textit{Or} operator for selection, and then press the backspace key to delete it.
The \textit{And} operator is added by selecting \textit{Add} in the drop-down box and clicking on the \Add button.
By linking the new operator to its dependency operators and the output operator, users can easily fix the structural error.

\begin{figure}[!t]
\centering
\includegraphics[width=1\linewidth]{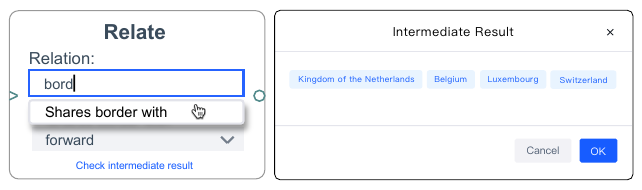}
    \caption{
    Left: Screenshot of the auto-completion in slot-filling. 
    Right: Screenshot of the intermediate result of the \textit{And} operator, which shows the satisfied countries.
    }
    \label{fig:complete}
\end{figure}

The other issue is the falsely recognized argument for the \textit{Relate} operator. 
The desired countries should share borders with \textit{Germany}, rather than the \textit{statement is subject of} relation.
The relation name specified by the KB schema is auto-completed in the pop-up drop down box, as shown in Figure~\ref{fig:complete}.

The intermediate result of each knowledge operator is a powerful tool to diagnose the KoPL program.
It also serves as an interpretation to the question's answer.
By expanding the intermediate result of the \textit{And} operator, as shown in the right part of Figure~\ref{fig:complete}, we are able to know which countries are taken into account.

\subsection{KoPL Engine}

The high performing KoPL engine incorporated in the back-end is developed as an independent extension for Python. 
It provides one line installation code from the command line by running ``\texttt{pip install kopl-engine}''.
Users can execute the KoPL program using the scripts provided at the end of this section.

Users are first required to provide the KB in JSON file per the request by KoPL\footnote{\href{kopl.xlore.cn/en/doc/4\_helloworld.html}{\url{https://kopl.xlore.cn/en/doc/4\_helloworld}}}.
The execution engine is initialized by converting the KB into data structure in the memory and constructing all the indices.
Before executing the KoPL program, the engine parses the program represented in Python data structure (See Appendix~\ref{sec:app:data} for the data structure introduction.) into the data structure used inside the engine.
After that, users can call the \texttt{forward} method of the engine to get execution results.

\begin{lstlisting}[language=Python, label={code:kopl}]
from kopl_engine import engine
# Knowledge base preparation
kb = engine.init("kb.json")
# Data structure conversion
p = engine.parse_program(program)
# Program execution with 
# intermediate result tracing
result = engine.forward(
         kb, p, trace=True)
\end{lstlisting}






\section{Evaluation}

We evaluate the execution efficiency of the back-end KoPL engine.
We also perform user study and case study to examine the limitations of VisKoP.

\subsection{Efficiency}

\textbf{KB preparation.}
VisKoP is deployed on a million-entity-level KB extracted from Wikidata.
In particular, we use the original Wikdiata dump\footnote{\url{https://dumps.wikimedia.org/wikidatawiki/entities/latest-all.json.bz2}} and only keep the entities that have a Wikipedia page.
The statistics is shown in Table~\ref{tab:wikidata}. 

\begin{table}[!ht]
\centering
\scalebox{0.88}{
\begin{tabular}{cccccc}
\toprule
\# Entity  & \# Concept & \# Relation & \# Attribute\\
\midrule
6,284,269 & 68,261 & 1,080 & 1,352 \\
\bottomrule
\end{tabular}
}
\caption{Statistics of the knowledge base.}
\label{tab:wikidata}
\end{table}


\textbf{Experimental setup.}
We use the training data of KQA Pro~\citep{kqapro2022kopl} as the test-bed, which contains 94,376 quries in both KoPL and SPARQL program.
We compare VisKoP against the original KoPL engine released by \citet{kqapro2022kopl}.
We also compare it with Virtuoso for the SPARQL queries.
All experiments are conducted on a single Intel Xeon 5218R CPU with $1.0$TB RAM.
We use wall time as the comparison metric.

\begin{table}[!thbp]
\centering
\scalebox{0.88}{
\begin{tabular}{c|rrrrrrrrrr}
\toprule
Engine     & \begin{tabular}{c}VisKoP\end{tabular} & \begin{tabular}{c}KoPL\end{tabular} & \begin{tabular}{c}Virtuoso\end{tabular} \\
\midrule
Wall Time    & 111.5 ms & 1775.8 ms & 535.1 ms \\
\bottomrule
\end{tabular}
}
\caption{Running time averaged over all the queries.}
\label{tab:efficiency}
\end{table}

\begin{figure}[!th]
\centering
\includegraphics[width=0.93\linewidth]{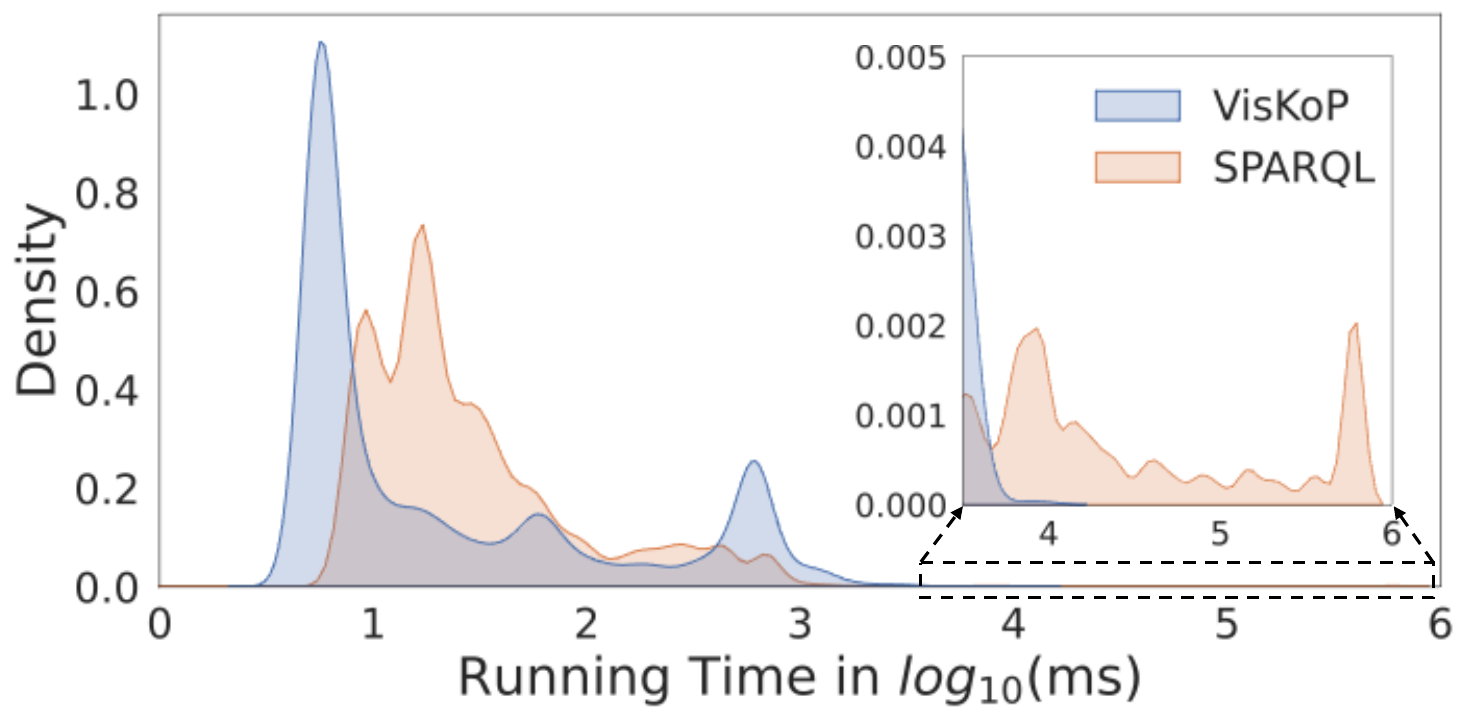}
    \vspace{-0.05in}
    \caption{
    Running time distribution.
    }
    \label{fig:time}
\end{figure}

The averaged running time is reported in Table~\ref{tab:efficiency}.
VisKoP is almost $16\times$ faster than the original KoPL engine and $5\times$ faster than Virtuoso executing equivalent SPARQL queries.
We also show the running time distribution of VisKoP and Virtuoso in Figure~\ref{fig:time}.
VisKoP is faster than Virtuoso because: 
(1) The distribution peak of VisKoP comes smaller than Virtuoso;
(2) The maximum running time of VisKoP is much smaller than Virtuoso.









\subsection{User Study and Case Study}\label{sec:userstudy}

We manually annotate $20$ natural language questions which cannot be correctly answered without user correction and ask $6$ different users to use VisKoP to find the answer.
After users interact with VisKoP, the accuracy rate reaches $65.8\%$, with an average of $110.7$ seconds per question and a median of $68.0$ seconds.
These results indicate that integrating human into the loop significantly broadens the boundaries of the KBQA system's capabilities.
Meanwhile, apart from knowledge elements not included in the KB, there are still questions that are extremely difficult to answer due to their obscure knowledge elements.
For example, to answer \textit{``How many video game is SONY published in 2020?''}, one need to find the \textit{Sony Interactive Entertainment} entity rather than the \textit{Sony}, which also occurs in the KB and our testers can hardly find the \textit{Sony Interactive Entertainment} entity.

\section{Related Works}
In general, KBQA methods can be grouped into two categories: 1) semantic parsing~\cite{emnlp13,querygraph,Cheng2017LearningSN,Liang2017NeuralSM,NPI,cao2022program}, which translates natural language questions into logical forms, whose execution on the KB achieves the answer; 2) information retrieval~\cite{emnlp14,acl14,kvmem,transfernet,Zhang2022SubgraphRE}, which ranks the entities from the retrieved question-specific sub-KB to get the answer. Our VisKoP falls into the semantic parsing category. Specifically, VisKoP translates a question into the multi-step program, pertaining to the neural program induction (NPI) paradigm~\cite{Lake2015HumanlevelCL,Neelakantan2017LearningAN,Liang2017NeuralSM,Wong2021LeveragingLT,kqapro2022kopl}.

The main challenge of NPI is that question-program parallel data are expensive to obtain and the program's huge search space makes the learning challenging. 
Existing works tackle this issue only by learning from question-answer pairs with various reinforcement learning techniques~\cite{Liang2017NeuralSM,saha2019complex} or synthesizing question-program data to alleviate the data scarcity problem~\cite{kqapro2022kopl,gu2021beyond}. 
In this paper, our VisKoP proposes a different solution to this task by integrating humans into the program induction loop, providing external human guidance to program induction model, and potentially improving the system robustness.

Compared with other KBQA systems, including ReTraCk~\cite{chen2021retrack}, SEMPRE~\cite{emnlp13}, TRANX~\cite{Yin2018TRANXAT}, DTQA~\cite{ibrahim2021dtqa}, our VisKoP is the first to enable users to interact with the system via a visual platform and intermediate results checking.






\section{Conclusion and Future Work}

We demonstrate VisKoP, a KBQA platform that allows users to monitor, edit, and debug KB queries.
VisKoP is also accompanied with a highly efficient engine that scales KoPL execution to a million-entity-level KB.
In the future, it is intriguing to allow users to customize the KB.
It is also important to provide guidance for users to recognize the true knowledge elements in the large scale KB.

\section*{Limitations}

As a KBQA system, VisKoP is still highly dependent on the correctness and broad knowledge coverage of the background KB.
It is extremely difficult to find the correct answer when the relevant knowledge elements are unincluded or incorrect in the KB.
Also, if there are confusing knowledge elements, as we mention in Section~\ref{sec:userstudy} that users can hardly identify the \textit{Sony Interactive Entertainment} entity, it is difficult for users to correct the KoPL program.

\section*{Ethics Statement}

\paragraph{Intended Use.}
VisKoP is designed for users to edit their knowledge base queries with graphical elements.

\paragraph{Potential Misuse.}
As we count, there are  $339,531$ human female entities and $1,458,903$ male entities in total.
It can lead to gender biased answers on the grounds that a number of females do not exist in the KB.
This problem stems from the imbalanced data (Wikidata), and can be solved when Wikidata includes more female entities.
Therefore, it's important to allow users to debug the knowledge base in future work.

\paragraph{Data.}
The VisKoP is built on a high-quality subset of Wikidata, which attributes to the intelligence of the crowd.

\paragraph{User Study.}
The participants in the user study part are volunteers recruited from graduate students majoring in engineering.
Before the user study experiments, all participants are provided with detailed guidance in both written and oral form.
The only recorded user-related information is usernames, which are anonymized and used as identifiers to mark different participants.

\section*{Acknowledgments}
This work is supported by National Key R\&D Program of China (2020AAA0105203), and a grant from the Institute for Guo Qiang, Tsinghua University (2019GQB0003)

\bibliography{refine}

\begin{thebibliography}{30}
\expandafter\ifx\csname natexlab\endcsname\relax\def\natexlab#1{#1}\fi

\bibitem[{Abdelaziz et~al.(2021)Abdelaziz, Ravishankar, Kapanipathi, Roukos,
  and Gray}]{ibrahim2021dtqa}
Ibrahim Abdelaziz, Srinivas Ravishankar, Pavan Kapanipathi, Salim Roukos, and
  Alexander~G. Gray. 2021.
\newblock \href {https://ojs.aaai.org/index.php/AAAI/article/view/17988} {A
  semantic parsing and reasoning-based approach to knowledge base question
  answering}.
\newblock In \emph{{AAAI}}.

\bibitem[{Ansari et~al.(2019)Ansari, Saha, Kumar, Bhambhani, Sankaranarayanan,
  and Chakrabarti}]{NPI}
Ghulam~Ahmed Ansari, Amrita Saha, Vishwajeet Kumar, Mohan Bhambhani, Karthik
  Sankaranarayanan, and Soumen Chakrabarti. 2019.
\newblock \href {https://doi.org/10.24963/ijcai.2019/679} {Neural program
  induction for {KBQA} without gold programs or query annotations}.
\newblock In \emph{Proceedings of the Twenty-Eighth International Joint
  Conference on Artificial Intelligence, {IJCAI} 2019, Macao, China, August
  10-16, 2019}, pages 4890--4896. ijcai.org.

\bibitem[{Berant et~al.(2013)Berant, Chou, Frostig, and Liang}]{emnlp13}
Jonathan Berant, Andrew Chou, Roy Frostig, and Percy Liang. 2013.
\newblock \href {https://aclanthology.org/D13-1160} {Semantic parsing on
  {F}reebase from question-answer pairs}.
\newblock In \emph{Proceedings of the 2013 Conference on Empirical Methods in
  Natural Language Processing}, pages 1533--1544, Seattle, Washington, USA.
  Association for Computational Linguistics.

\bibitem[{Bordes et~al.(2014)Bordes, Chopra, and Weston}]{emnlp14}
Antoine Bordes, Sumit Chopra, and Jason Weston. 2014.
\newblock \href {https://doi.org/10.3115/v1/D14-1067} {Question answering with
  subgraph embeddings}.
\newblock In \emph{Proceedings of the 2014 Conference on Empirical Methods in
  Natural Language Processing ({EMNLP})}, pages 615--620, Doha, Qatar.
  Association for Computational Linguistics.

\bibitem[{Cao et~al.(2022{\natexlab{a}})Cao, Shi, Pan, Nie, Xiang, Hou, Li, He,
  and Zhang}]{kqapro2022kopl}
Shulin Cao, Jiaxin Shi, Liangming Pan, Lunyiu Nie, Yutong Xiang, Lei Hou,
  Juanzi Li, Bin He, and Hanwang Zhang. 2022{\natexlab{a}}.
\newblock \href {https://aclanthology.org/2022.acl-long.422} {{KQA} pro: A
  dataset with explicit compositional programs for complex question answering
  over knowledge base}.
\newblock In \emph{ACL}.

\bibitem[{Cao et~al.(2022{\natexlab{b}})Cao, Shi, Yao, Lv, Yu, Hou, Li, Liu,
  and Xiao}]{cao2022program}
Shulin Cao, Jiaxin Shi, Zijun Yao, Xin Lv, Jifan Yu, Lei Hou, Juanzi Li,
  Zhiyuan Liu, and Jinghui Xiao. 2022{\natexlab{b}}.
\newblock \href {https://aclanthology.org/2022.acl-long.559} {Program transfer
  for answering complex questions over knowledge bases}.
\newblock In \emph{ACL}.

\bibitem[{Chen et~al.(2021)Chen, Liu, Yu, Lin, Lou, and
  Jiang}]{chen2021retrack}
Shuang Chen, Qian Liu, Zhiwei Yu, Chin-Yew Lin, Jian-Guang Lou, and Feng Jiang.
  2021.
\newblock \href {https://doi.org/10.18653/v1/2021.acl-demo.39} {{R}e{T}ra{C}k:
  A flexible and efficient framework for knowledge base question answering}.
\newblock In \emph{Proceedings of the 59th Annual Meeting of the Association
  for Computational Linguistics and the 11th International Joint Conference on
  Natural Language Processing: System Demonstrations}, pages 325--336, Online.
  Association for Computational Linguistics.

\bibitem[{Cheng et~al.(2017)Cheng, Reddy, Saraswat, and
  Lapata}]{Cheng2017LearningSN}
Jianpeng Cheng, Siva Reddy, Vijay Saraswat, and Mirella Lapata. 2017.
\newblock \href {https://doi.org/10.18653/v1/P17-1005} {Learning structured
  natural language representations for semantic parsing}.
\newblock In \emph{Proceedings of the 55th Annual Meeting of the Association
  for Computational Linguistics (Volume 1: Long Papers)}, pages 44--55,
  Vancouver, Canada. Association for Computational Linguistics.

\bibitem[{Cui et~al.(2016)Cui, Xiao, and Wang}]{wanyun2016kbqafreebase}
Wanyun Cui, Yanghua Xiao, and Wei Wang. 2016.
\newblock \href {http://www.ijcai.org/Abstract/16/640} {{KBQA:} an online
  template based question answering system over freebase}.
\newblock In \emph{Proceedings of the Twenty-Fifth International Joint
  Conference on Artificial Intelligence, {IJCAI} 2016, New York, NY, USA, 9-15
  July 2016}, pages 4240--4241. {IJCAI/AAAI} Press.

\bibitem[{Gu et~al.(2021)Gu, Kase, Vanni, Sadler, Liang, Yan, and
  Su}]{gu2021beyond}
Yu~Gu, Sue Kase, Michelle Vanni, Brian Sadler, Percy Liang, Xifeng Yan, and
  Yu~Su. 2021.
\newblock \href {https://dl.acm.org/doi/abs/10.1145/3442381.3449992} {Beyond
  iid: three levels of generalization for question answering on knowledge
  bases}.
\newblock In \emph{WWW'21}.

\bibitem[{H{\"o}ffner et~al.(2013)H{\"o}ffner, Unger, B{\"u}hmann, Lehmann,
  Ngomo, Gerber, and Cimiano}]{hoffner2013tbsl}
K~H{\"o}ffner, C~Unger, L~B{\"u}hmann, J~Lehmann, ACN Ngomo, D~Gerber, and
  P~Cimiano. 2013.
\newblock \href
  {https://www.academia.edu/19032334/TBSL_Question_Answering_System_Demo} {Tbsl
  question answering system demo}.
\newblock In \emph{Proceedings of the 4th Conference on Knowledge Engineering
  Semantic Web}.

\bibitem[{Lake et~al.(2015)Lake, Salakhutdinov, and
  Tenenbaum}]{Lake2015HumanlevelCL}
B.~Lake, R.~Salakhutdinov, and J.~Tenenbaum. 2015.
\newblock \href
  {https://www.cs.cmu.edu/~rsalakhu/papers/LakeEtAl2015Science.pdf}
  {Human-level concept learning through probabilistic program induction}.
\newblock \emph{Science}, 350:1332 -- 1338.

\bibitem[{Lan et~al.(2021)Lan, He, Jiang, Jiang, Zhao, and rong
  Wen}]{Lan2021ASO}
Yunshi Lan, Gaole He, Jinhao Jiang, Jing Jiang, Wayne~Xin Zhao, and Ji~rong
  Wen. 2021.
\newblock \href {https://www.ijcai.org/proceedings/2021/0611.pdf} {A survey on
  complex knowledge base question answering: Methods, challenges and
  solutions}.
\newblock In \emph{IJCAI}.

\bibitem[{Lewis et~al.(2020)Lewis, Liu, Goyal, Ghazvininejad, Mohamed, Levy,
  Stoyanov, and Zettlemoyer}]{lewis2020bart}
Mike Lewis, Yinhan Liu, Naman Goyal, Marjan Ghazvininejad, Abdelrahman Mohamed,
  Omer Levy, Veselin Stoyanov, and Luke Zettlemoyer. 2020.
\newblock \href {https://doi.org/10.18653/v1/2020.acl-main.703} {{BART}:
  Denoising sequence-to-sequence pre-training for natural language generation,
  translation, and comprehension}.
\newblock In \emph{Proceedings of the 58th Annual Meeting of the Association
  for Computational Linguistics}, pages 7871--7880, Online. Association for
  Computational Linguistics.

\bibitem[{Liang et~al.(2017)Liang, Berant, Le, Forbus, and
  Lao}]{Liang2017NeuralSM}
Chen Liang, Jonathan Berant, Quoc Le, Kenneth~D. Forbus, and Ni~Lao. 2017.
\newblock \href {https://doi.org/10.18653/v1/P17-1003} {Neural symbolic
  machines: Learning semantic parsers on {F}reebase with weak supervision}.
\newblock In \emph{Proceedings of the 55th Annual Meeting of the Association
  for Computational Linguistics (Volume 1: Long Papers)}, pages 23--33,
  Vancouver, Canada. Association for Computational Linguistics.

\bibitem[{Mihindukulasooriya et~al.(2020)Mihindukulasooriya, Rossiello,
  Kapanipathi, Abdelaziz, Ravishankar, Yu, Gliozzo, Roukos, and
  Gray}]{mihindukulasooriya2020leveraging}
Nandana Mihindukulasooriya, Gaetano Rossiello, Pavan Kapanipathi, Ibrahim
  Abdelaziz, Srinivas Ravishankar, Mo~Yu, Alfio Gliozzo, Salim Roukos, and
  Alexander Gray. 2020.
\newblock \href
  {https://link.springer.com/chapter/10.1007/978-3-030-62419-4_23} {Leveraging
  semantic parsing for relation linking over knowledge bases}.
\newblock In \emph{ISWC}.

\bibitem[{Miller et~al.(2016)Miller, Fisch, Dodge, Karimi, Bordes, and
  Weston}]{kvmem}
Alexander Miller, Adam Fisch, Jesse Dodge, Amir-Hossein Karimi, Antoine Bordes,
  and Jason Weston. 2016.
\newblock \href {https://doi.org/10.18653/v1/D16-1147} {Key-value memory
  networks for directly reading documents}.
\newblock In \emph{Proceedings of the 2016 Conference on Empirical Methods in
  Natural Language Processing}, pages 1400--1409, Austin, Texas. Association
  for Computational Linguistics.

\bibitem[{Neelakantan et~al.(2017)Neelakantan, Le, Abadi, McCallum, and
  Amodei}]{Neelakantan2017LearningAN}
Arvind Neelakantan, Quoc~V. Le, Mart{\'{\i}}n Abadi, Andrew McCallum, and Dario
  Amodei. 2017.
\newblock \href {https://openreview.net/forum?id=ry2YOrcge} {Learning a natural
  language interface with neural programmer}.
\newblock In \emph{5th International Conference on Learning Representations,
  {ICLR} 2017, Toulon, France, April 24-26, 2017, Conference Track
  Proceedings}. OpenReview.net.

\bibitem[{Nie et~al.(2022)Nie, Cao, Shi, Sun, Tian, Hou, Li, and
  Zhai}]{graphqir}
Lunyiu Nie, Shulin Cao, Jiaxin Shi, Jiuding Sun, Qi~Tian, Lei Hou, Juanzi Li,
  and Jidong Zhai. 2022.
\newblock \href {https://aclanthology.org/2022.emnlp-main.394} {{G}raph{Q}
  {IR}: Unifying the semantic parsing of graph query languages with one
  intermediate representation}.
\newblock In \emph{EMNLP}.

\bibitem[{Saha et~al.(2019)Saha, Ansari, Laddha, Sankaranarayanan, and
  Chakrabarti}]{saha2019complex}
Amrita Saha, Ghulam~Ahmed Ansari, Abhishek Laddha, Karthik Sankaranarayanan,
  and Soumen Chakrabarti. 2019.
\newblock \href {https://doi.org/10.1162/tacl_a_00262} {Complex program
  induction for querying knowledge bases in the absence of gold programs}.
\newblock \emph{Transactions of the Association for Computational Linguistics},
  7:185--200.

\bibitem[{Shaw et~al.(2021)Shaw, Chang, Pasupat, and
  Toutanova}]{shaw2021compositional}
Peter Shaw, Ming-Wei Chang, Panupong Pasupat, and Kristina Toutanova. 2021.
\newblock \href {https://doi.org/10.18653/v1/2021.acl-long.75} {Compositional
  generalization and natural language variation: Can a semantic parsing
  approach handle both?}
\newblock In \emph{Proceedings of the 59th Annual Meeting of the Association
  for Computational Linguistics and the 11th International Joint Conference on
  Natural Language Processing (Volume 1: Long Papers)}, pages 922--938, Online.
  Association for Computational Linguistics.

\bibitem[{Shi et~al.(2021)Shi, Cao, Hou, Li, and Zhang}]{transfernet}
Jiaxin Shi, Shulin Cao, Lei Hou, Juanzi Li, and Hanwang Zhang. 2021.
\newblock \href {https://doi.org/10.18653/v1/2021.emnlp-main.341}
  {{T}ransfer{N}et: An effective and transparent framework for multi-hop
  question answering over relation graph}.
\newblock In \emph{Proceedings of the 2021 Conference on Empirical Methods in
  Natural Language Processing}, pages 4149--4158, Online and Punta Cana,
  Dominican Republic. Association for Computational Linguistics.

\bibitem[{Shin et~al.(2021)Shin, Lin, Thomson, Chen, Roy, Platanios, Pauls,
  Klein, Eisner, and Van~Durme}]{Shin2021ConstrainedLM}
Richard Shin, Christopher Lin, Sam Thomson, Charles Chen, Subhro Roy,
  Emmanouil~Antonios Platanios, Adam Pauls, Dan Klein, Jason Eisner, and
  Benjamin Van~Durme. 2021.
\newblock \href {https://doi.org/10.18653/v1/2021.emnlp-main.608} {Constrained
  language models yield few-shot semantic parsers}.
\newblock In \emph{Proceedings of the 2021 Conference on Empirical Methods in
  Natural Language Processing}, pages 7699--7715, Online and Punta Cana,
  Dominican Republic. Association for Computational Linguistics.

\bibitem[{Wang et~al.(2015)Wang, Berant, and Liang}]{Wang2015BuildingAS}
Yushi Wang, Jonathan Berant, and Percy Liang. 2015.
\newblock \href {https://doi.org/10.3115/v1/P15-1129} {Building a semantic
  parser overnight}.
\newblock In \emph{Proceedings of the 53rd Annual Meeting of the Association
  for Computational Linguistics and the 7th International Joint Conference on
  Natural Language Processing (Volume 1: Long Papers)}, pages 1332--1342,
  Beijing, China. Association for Computational Linguistics.

\bibitem[{Wong et~al.(2021)Wong, Ellis, Tenenbaum, and
  Andreas}]{Wong2021LeveragingLT}
Catherine Wong, Kevin Ellis, Joshua~B. Tenenbaum, and Jacob Andreas. 2021.
\newblock \href {http://proceedings.mlr.press/v139/wong21a.html} {Leveraging
  language to learn program abstractions and search heuristics}.
\newblock In \emph{Proceedings of the 38th International Conference on Machine
  Learning, {ICML} 2021, 18-24 July 2021, Virtual Event}, volume 139 of
  \emph{Proceedings of Machine Learning Research}, pages 11193--11204. {PMLR}.

\bibitem[{Xu et~al.(2016)Xu, Reddy, Feng, Huang, and Zhao}]{acl14}
Kun Xu, Siva Reddy, Yansong Feng, Songfang Huang, and Dongyan Zhao. 2016.
\newblock \href {https://doi.org/10.18653/v1/P16-1220} {Question answering on
  {F}reebase via relation extraction and textual evidence}.
\newblock In \emph{Proceedings of the 54th Annual Meeting of the Association
  for Computational Linguistics (Volume 1: Long Papers)}, pages 2326--2336,
  Berlin, Germany. Association for Computational Linguistics.

\bibitem[{Yih et~al.(2015)Yih, Chang, He, and Gao}]{querygraph}
Wen-tau Yih, Ming-Wei Chang, Xiaodong He, and Jianfeng Gao. 2015.
\newblock \href {https://doi.org/10.3115/v1/P15-1128} {Semantic parsing via
  staged query graph generation: Question answering with knowledge base}.
\newblock In \emph{Proceedings of the 53rd Annual Meeting of the Association
  for Computational Linguistics and the 7th International Joint Conference on
  Natural Language Processing (Volume 1: Long Papers)}, pages 1321--1331,
  Beijing, China. Association for Computational Linguistics.

\bibitem[{Yin et~al.(2021)Yin, Fang, Neubig, Pauls, Platanios, Su, Thomson, and
  Andreas}]{yin2021compositional}
Pengcheng Yin, Hao Fang, Graham Neubig, Adam Pauls, Emmanouil~Antonios
  Platanios, Yu~Su, Sam Thomson, and Jacob Andreas. 2021.
\newblock \href {https://doi.org/10.18653/v1/2021.naacl-main.225}
  {Compositional generalization for neural semantic parsing via span-level
  supervised attention}.
\newblock In \emph{Proceedings of the 2021 Conference of the North American
  Chapter of the Association for Computational Linguistics: Human Language
  Technologies}, pages 2810--2823, Online. Association for Computational
  Linguistics.

\bibitem[{Yin and Neubig(2018)}]{Yin2018TRANXAT}
Pengcheng Yin and Graham Neubig. 2018.
\newblock \href {https://doi.org/10.18653/v1/D18-2002} {{TRANX}: A
  transition-based neural abstract syntax parser for semantic parsing and code
  generation}.
\newblock In \emph{Proceedings of the 2018 Conference on Empirical Methods in
  Natural Language Processing: System Demonstrations}, pages 7--12, Brussels,
  Belgium. Association for Computational Linguistics.

\bibitem[{Zhang et~al.(2022)Zhang, Zhang, Yu, Tang, Tang, Li, and yan
  Chen}]{Zhang2022SubgraphRE}
Jing Zhang, Xiaokang Zhang, Jifan Yu, Jian Tang, Jie Tang, Cuiping Li, and Hong
  yan Chen. 2022.
\newblock \href {https://aclanthology.org/2022.acl-long.396/} {Subgraph
  retrieval enhanced model for multi-hop knowledge base question answering}.
\newblock In \emph{ACL}.

\end{thebibliography}
\bibliographystyle{acl_natbib}

\clearpage

\appendix

\section{KoPL Definition}\label{sec:app:kopl}

The functions used in this paper are the same as those mentioned in~\citet{kqapro2022kopl}, so we will not devote a great deal of space for details. The specific meaning of each function can be found in~\cite{kqapro2022kopl} or on our website~\footnote{\url{https://kopl.xlore.cn/en/doc/2_function.html}}.
Here we only briefly introduce the philosophy of these operators:

\noindent\textbf{Query Operators}
find and return the knowledge elements in the KB by matching their names. \eg\ \textit{Find} returns the corresponding entities according to the input entity name.

\noindent\textbf{Filter Operators}
take a set of knowledge elements as input, and keep the knowledge elements that satisfy the given conditions as output. \eg\ \textit{FilterConcept} takes a set of entities as input and output entities that belong to a given concept.

\noindent\textbf{Verification Operators}
are used to determine whether the output of the previous function has some relationship to the given value. This type of operators is often used to answer judgement questions. \eg\ \textit{VerifyNum} can judge whether the function output is greater than (less than, equal to) a given value.

\noindent\textbf{Selection Operators} select some knowledge elements from the output of previous function under the given condition. \eg\ \textit{SelectAmong} can select the entity with the largest or smallest value of an attribute from a given set.

\noindent\textbf{Set Operators} do inter-set operations on the output of two functions. \eg\ \textit{And} can take the union of two sets.

\section{KoPL Program Format}\label{sec:app:data}

In python, each knowledge operator is represented as a \texttt{Dict} in Python with three keys:
\texttt{function} corresponds to the name of the knowledge operator.
\texttt{inputs} corresponds to the argument inputs of the knowledge operator.
And \texttt{dependencies} corresponds to the dependency inputs of the knowledge operator.
For example, KoPL program in Figure~\ref{fig:screen} can be represented as:

\phantom{a} \\ \phantom{a} \\ \phantom{a} \\ \phantom{a} \\
\phantom{a} \\ \phantom{a} \\ 

\begin{lstlisting}[language=Python, label={code:program}]
program = {[
  {
    "function": "Find",
    "inputs": ["France"], 
    "dependencies":[-1,-1]
  },
  {
    "function": "Relate",
    "inputs": ["shares border with", "backward"],
    "dependencies":[0]
  },
  {
    "function": "FilterConcept",
    "inputs": ["country"], 
    "dependencies": 
    [1]
  },
  {
    "function": "Find", 
    "inputs": ["Germany"], 
    "dependencies":[]
  },
  {
    "function": "Relate", 
    "inputs": ["statement is subject of","forward"],
    "dependencies": [3]
  },
  {
    "function": "FilterConcept", 
    "inputs": ["country"], 
    "dependencies": [4]
  },
  {
    "function": "Or",
    "inputs": [],
    "dependencies": [2,5]
  },
  {
    "function": "Count",
    "inputs": [],
    "dependencies":[6]
  }
]}
\end{lstlisting}

\section{Questions for User Study}\label{sec:app:question}

We list the 20 questions used in the user study and the corresponding answers in Table~\ref{tab:20_questions}.

\begin{table*}[hbt!]
  \centering
  \hspace*{-0.32cm}
  \setlength\tabcolsep{5.5pt}
  \scalebox{0.8}{
  \begin{tabular}{lr}
  \toprule
    \textbf{Question} & \textbf{Correct Answer} \\ 
  \midrule
  How many Olympic Games has LeBron James competed in? & 3 \\
  What is the name of the company that makes the game "The Legend of Zelda"? & Nintendo\\
  How many teams have both LeBron and Kobe played for? & 1 \\
  Is China more than 9.7 million square kilometres in size? & No \\
  Which is the largest province in China by area? & Qinghai \\
  How many countries are there in the European Union? & 27 \\
  How many times has Federer won a tennis competition? & 29 \\
  Which country is Google headquartered in? & United States of America \\
  How many international airports are there in Germany? & 15 \\
  Who is lighter, the iPhone X or the Samsung S10? & Samsung Galaxy S10 \\
  Which is the highest of all the mountains in South America and Africa? & Aconcagua \\
  How many video game is SONY published in 2020? & 566 \\
  Who is the next president after Barack Obama? & Donald Trump \\
  At which college did Geoffrey Hinton get his degree? & University of Edinburgh \\
  How many people have won the Nobel Prize in Physics? & 186 \\
  What award did Lawrence win for The Hunger Games? & MTV Movie Award for Best Female Performance \\
  How many states does the United States contain? & 50 \\
  Which is the highest mountain in Asia? & Mount Everest \\
  What year was the team owned by Jordan founded? & 1988 \\
  In which country is Nikon headquartered? & Japan \\
  \bottomrule
  \end{tabular}
  }
  \vspace{-0.1cm}
  \caption{20 questions used for user study.}
  \label{tab:20_questions}
  \vspace{-0.2cm}
\end{table*}

\end{document}